\documentclass[runningheads]{llncs}
\usepackage[T1]{fontenc}
\usepackage{graphicx,verbatim}
\usepackage{booktabs}
\usepackage[table]{xcolor}
\usepackage{subcaption}
\usepackage{marvosym}
\usepackage{color}
\newcommand\UrlFont{\color{blue}\rmfamily}
\begin{document}
\title{MedPAO-Assist: Edge-Deployable, Modality-Aware Agentic AI for Radiology Report Structuring}
\titlerunning{MedPAO-Assist}

\author{Shrish Shrinath Vaidya\inst{1}(\Letter)\thanks{Corresponding author: \email{shrish.s.vaidya@dsai.iitm.ac.in}}\orcidID{0009-0006-5438-9500} \and
Minmini Selvam\inst{2} \and
Ganapathy Krishnamurthi\inst{1}}
\authorrunning{S. S. Vaidya et al.}
\institute{Department of Data Science and Artificial Intelligence,
Indian Institute of Technology Madras, Chennai, India \and
TANUH -- The AI Centre of Excellence in Healthcare, Indian Institute of Science, Bengaluru, India}

  
\maketitle

\begin{abstract}
The adoption of Large Language Models for radiology report structuring has been largely dominated by either proprietary cloud-based models or large open models requiring high-end GPUs for deployment, which is a critical bottleneck for clinical adoption in radiology workflows that demand
data privacy and offline operation. To address this, we present MedPAO-Assist, an edge-deployable agentic framework that generates structured radiology reports directly from radiologist dictation and voice commands, requiring only $\approx3.3$ GB peak RAM and achieving an average end-to-end latency of $\approx31$ seconds on a standard laptop CPU through an asynchronous, multi-threaded execution pipeline. The system currently supports four major imaging modalities -- Chest X-ray, Chest CT, Abdomen CT, and Brain MRI. To our knowledge, it is among the earliest agentic systems to run its entire pipeline with a Small Language Model (SLM) in the 0.6B-parameter range. This multi-agent architecture, built around an Extract-Ground-Generate paradigm, performs concept extraction, SNOMED-CT grounding, and structured report generation, followed by a speech-driven human-in-the-loop correction phase, together forming a fully "no-touch" reporting workflow that requires no keyboard input, external API calls, or internet connectivity. We evaluate each pipeline stage individually -- concept extraction, ontology grounding, and concept categorization and report hardware profiling results demonstrating the system's feasibility for real-time, CPU-only clinical deployment. The complete code and setup instructions are here: {\UrlFont https://github.com/MiRL-IITM/MedPAO-Assist}.


\end{abstract}

%
%
\section{Introduction}
Radiology reports are formal documents that institutions rely on for billing, quality auditing, and regulatory compliance \cite{durack2014value}. Professional societies now recommend structured reporting be integrated directly into radiologists' workflows \cite{european2018esr,CHUNG}. Yet manually structuring a report, or correcting an automatically generated one, remains time-consuming on top of an already demanding clinical workload \cite{baileyunderstanding}, motivating systems that perform accurate, protocol-aware report structuring directly from dictation without disrupting existing workflow.



Automated structured reporting first emerged through rule-based methods \cite{kreimeyer2017natural} and traditional ML models \cite{steinkamp2019toward}, both of which depend on heavy manual annotation and generalize poorly beyond the templates they were built for. Building on this, an early speech-driven NLP dialogue system demonstrated automated structuring for abdominal CT in suspected urolithiasis \cite{jorg2023efficient}, but, like its rule-based predecessors, remained tied to a single fixed protocol rather than a modality-agnostic design. Commercial products such as MedicAI \cite{medicai2026}, RadAI \cite{radai2026}, and Quillr \cite{quillr2026} have since brought speech-driven structuring into clinical use at scale, yet all three are exclusively cloud-hosted, with no on-premise deployment option -- a critical gap for institutions where patient dictations and imaging-derived data cannot leave the local clinical environment. More recently, LLM-based approaches \cite{li2024promptguided,thyroid2024chatgpt,busch2025large} have shown that GPT-3.5/4 can structure reports well given sufficient clinical context, and RadExtract \cite{goel_langextract} similarly applies Google's LangExtract library with Gemini to the task; however, these approaches rely on proprietary, cloud-hosted models, raising data privacy and hospital-compliance concerns. Efforts more attuned to deployment constraints have instead trained
dedicated pipelines for CXR-specific structuring \cite{pellegrini2026protosr} or lightweight, edge-oriented models \cite{moll2025structuring}; the latter shows that sub-300M-parameter models can rival LLMs in performance, but still requires substantial supervised training data and remains confined to a single modality. Across this entire trajectory, methods are further framed as fully automated pipelines rather than human-AI co-adaptive tools, offering no structured, memory-informed pathway for a radiologist to correct and refine model output as part of the reporting workflow itself.

Beyond the workflow gaps above, realizing such a system safely is constrained by two further factors: patient dictations constitute protected health information, making cloud-hosted large language models (LLMs) a poor fit for real-time clinical deployment, while on-device Small Language Models (SLMs) that avoid this risk are known to suffer degraded reasoning on long-context inputs \cite{lostinmiddle,ruler} -- directly at odds with the length of typical radiology dictations.

To address these gaps, we propose \textbf{MedPAO-Assist}, an agentic AI
framework with the following contributions:
\begin{itemize}
\item A \textbf{Multi-Agent framework} (MAS) for protocol-driven, multi-modality report structuring.
\item Efficient use of \textbf{SLMs and PEFT methods} for medical concept extraction and categorization.
\item \textbf{Low-resource CPU deployment} with minimal latency and memory footprint.
\item Integration of MAS with \textbf{ASR} to provide a multi-modality, protocol-compliant, \textbf{touch-free} reporting interface.
\end{itemize}


\section{Methodology}
\subsection{System Architecture}
\begin{figure}[!htbp]
    \centering
    \includegraphics[width=0.99\linewidth]{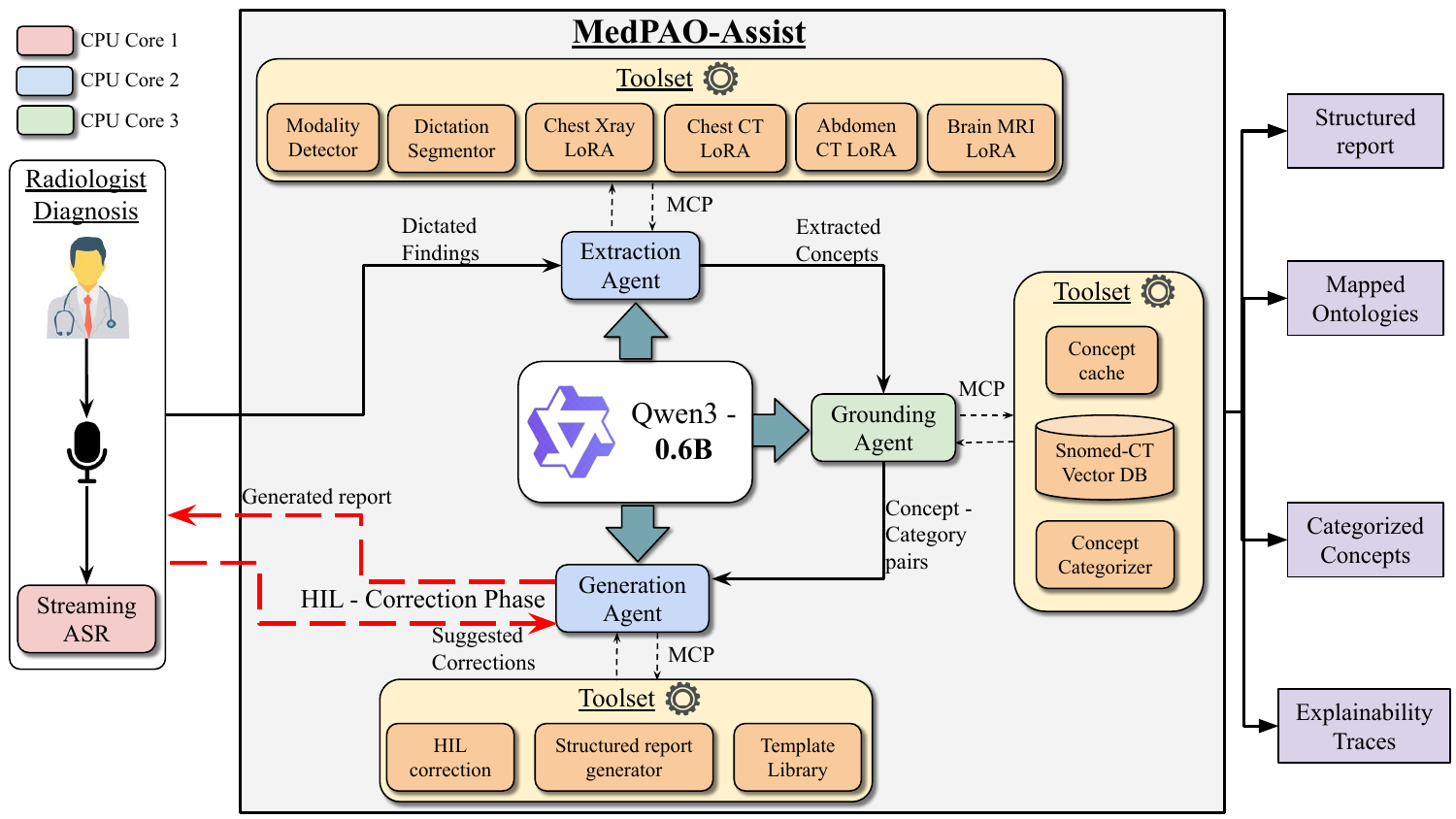}
    \caption{Proposed System Architecture}
    \label{fig:sys_arch}
\end{figure}

MedPAO-Assist employs a multi-threaded architecture that achieves low end-to-end latency by enabling real-time speech streaming through three parallel threads (Fig.~1). All agents share a Qwen3-0.6B Small Language Model (SLM)~\cite{qwen3} with task-specific LoRA adapters for specialized reasoning. Streaming transcription is performed using INT8-quantized MedASR~\cite{medasr}, while the SLM and LoRA adapters are deployed in GGUF format with FP16 inference on the target CPU. The system assumes a clinically familiar dictation workflow in which the radiologist states the imaging modality, followed by the findings and impression, allowing the Modality Detector to identify the appropriate protocol template early in the stream. Inspired by MedPAO~\cite{medpao}, the framework follows an \textbf{EGG (Extract--Ground--Generate)} paradigm, where the ASR, Extraction, and Grounding modules operate asynchronously on incoming speech segments to minimize latency. The Generation Agent then aggregates the extracted and grounded information to produce the final structured report, followed by a human-in-the-loop correction stage.

\subsection{Extraction Agent}
\label{sec:conc_extr}

Small Language Models (SLMs) exhibit degraded performance on long-context inputs~\cite{lostinmiddle,ruler}. To address this, the Extraction Agent identifies and retains only clinically relevant findings(i.e., concepts), transforming downstream processing from sentence-level to concept-level reasoning while preserving essential diagnostic information. The same SLM is used by the Modality Detector to identify the imaging modality and retrieve the corresponding protocol template. A rule-based Dictation Segmentor further separates the report into \textit{Findings} and \textit{Impression} sections using regular expressions, enabling independent downstream processing.

\subsubsection{Concept Extraction Tools:}
Separate LoRA adapters are trained for each imaging modality using a three-stage pipeline of Supervised Fine-Tuning (SFT), Knowledge Distillation(KD), and Instruction Tuning(IT)~\cite{deepseek-r1,minillm,qwen3}. First, the base model is LoRA fine-tuned on 500K MIMIC-IV-Note~\cite{mimic4-note} radiology reports to adapt it to radiology language. Next, concept extraction knowledge is distilled from Qwen3-32B, which performs zero-shot concept extraction on another 500K MIMIC-IV-Note reports, with its predictions and output logits used to continue training the same adapter. Finally, the adapter is instruction-tuned on the RadGraph-XL~\cite{radgraph-xl} training set, containing report--concept pairs from Chest X-ray, Chest CT, Abdomen CT, and Brain MRI.

\subsection{Grounding agent}
\label{sec:grounding_agent}

\begin{figure}[!htbp]
    \centering
    \includegraphics[trim=0cm 0cm 0cm 9.2cm, clip,width=1\linewidth]{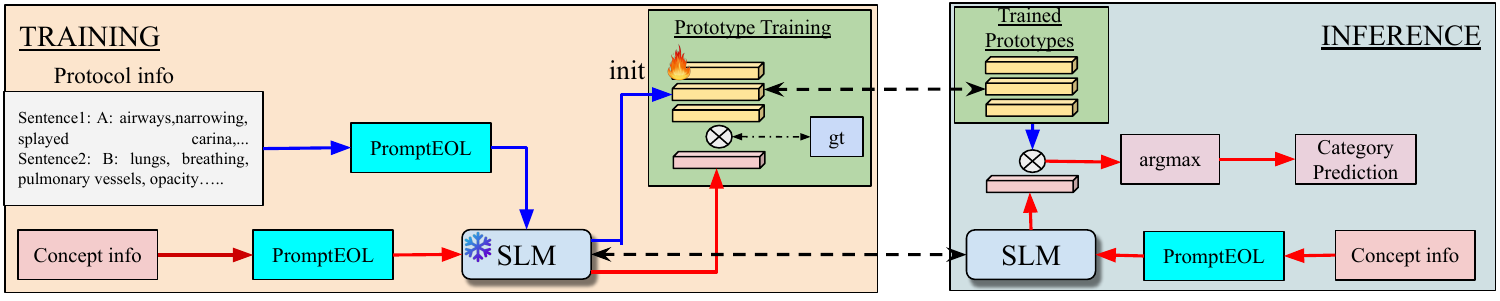}
    \caption{Proposed Strategy for concept categorization using PromptEOL\cite{prompteol}}
    \label{fig:conc_categorizer}
\end{figure}
\subsubsection{Ontology Mapping:} The Ontology Mapping module is responsible for grounding the extracted medical concepts in structured clinical knowledge. SNOMED-CT is adopted as the reference ontology due to its comprehensive coverage of radiological terminology. Extracted concepts are mapped to their semantically corresponding SNOMED-CT terms using SapBERT~\cite{sapbert} embeddings and embedding-based similarity search. To enable efficient CPU inference and reduce memory consumption, the similarity search employs IVF indexing and SQ8 scalar quantization of embeddings, as implemented in FAISS~\cite{faiss}. For each matched SNOMED-CT concept, its corresponding description and ancestor concepts are also retrieved to facilitate the subsequent categorization stage.

\subsubsection{Concept Categorization:} Unlike previous approaches~\cite{medpao}, our framework cannot rely on the reasoning capabilities of larger language models to directly assign extracted concepts to protocol-specific categories. Instead, we leverage the inherent sentence embedding capabilities of Small Language Models (SLMs), following the PromptEOL framework~\cite{prompteol}. Building upon this approach, we introduce a learnable embedding layer to optimize protocol prototypes(Fig.~\ref{fig:conc_categorizer}). Experimental evaluation demonstrates that the proposed method achieves strong performance with as few as 20-50 annotated concept--category training pairs, indicating that the SLM-derived prototype embeddings provide an effective initialization for low-resource learning. Protocol templates for the different imaging modalities were selected in consultation with a radiologist and based on the RSNA Reporting Template Library~\cite{rsna_radreport}. Synthetic concept--category pairs were subsequently generated using Claude Sonnet 5 according to the protocol definitions and used for training and evaluation of the proposed concept categorization framework.

\subsection{Generation agent}

\subsubsection{Structured Generation:} The Generation Agent is responsible for producing the final structured report by aggregating the extracted concepts, their corresponding source sentences, and the concept categorization results. As the source sentences remain in spoken dictation form (e.g., "I could see...", "I am looking at...") until this stage, they must be transformed into clinically appropriate reporting language. To achieve this, the structured report generation module employs a LoRA adapter over the base SLM to convert the dictated phrases into their corresponding medical report equivalents. This LoRA adapter is fine-tuned on a synthetic dataset generated using Claude Sonnet 5, GPT-OSS-120B, Llama-3.3-70B-Versatile, and DeepSeek V3.

\subsubsection{Human-in-Loop Correction:} Since the proposed framework operates using a 0.6B-parameter Small Language Model (SLM), occasional prediction errors are inevitable. To address this, MedPAO-Assist incorporates a human-in-the-loop correction stage following the initial structured report generation. During this phase, the ASR model and the base SLM operate synchronously to process verbal corrections provided by the clinician or radiologist. The spoken feedback is transcribed and used to update the generated report, enabling efficient post-generation refinement while preserving a speech-driven workflow.


\section{Results and Analysis}

\subsection{Concept Extraction}

\begin{table}[!htbp]
\centering
\caption{Evaluation of Concept Extraction task with various phases of finetuning.}
\label{tab:conc_extraction}
\fontsize{8}{9}\selectfont
\begin{tabular}{lccccccc}
\toprule
\textbf{Method}
    & \textbf{F1-score} $\uparrow$
    & \textbf{Precision} $\uparrow$
    & \textbf{Recall} $\uparrow$
    & \textbf{Hamming Loss} $\downarrow$\\
\midrule
Base Model
    & 0.08
    & 0.19
    & 0.06
    & 0.0048\\
+ SFT
    & 0.08
    & 0.15
    & 0.07
    & 0.0047 \\
+ KD
    & 0.55
    & 0.66
    & 0.50
    & 0.0026 \\
\rowcolor{blue!10}
+ IT
    & 0.78
    & 0.82
    & 0.76
    & 0.0012 \\
\bottomrule
\end{tabular}

\end{table}

The proposed concept extraction pipeline described in Section~\ref{sec:conc_extr} is evaluated on the RadGraph-XL~test set, comprising approximately 290 radiology reports spanning Chest X-ray, Chest CT, Abdomen CT, and Brain MRI modalities. Since concept extraction is performed using a generative Small Language Model (SLM), the predictions are evaluated using fuzzy string matching with a similarity threshold of 80. The resulting classification metrics across the different stages of fine-tuning are presented in Table~\ref{tab:conc_extraction}. The results demonstrate a consistent improvement in performance across successive training stages, with the final model achieving an F1-score of 0.78 achieving a boost of 0.7 as compared to the base model. This substantial improvement highlights the effectiveness of the proposed three-stage fine-tuning strategy.


\subsection{Ontology mapping}

\begin{table}[!htbp]
\centering
\caption{Evaluation of SNOMED CT grounding under various FAISS index configurations.}
\label{tab:ontology_map}
\fontsize{8}{9}\selectfont
\begin{tabular}{lcccc}
\toprule
\textbf{Method}
    & \textbf{Top1-Acc.} $\uparrow$
    & \textbf{Top3-Acc.} $\uparrow$
    & \textbf{Latency(in sec)} $\downarrow$
    & \textbf{Peak RAM(in MB)} $\downarrow$\\
\midrule
Baseline\footnotemark[1]
    & 56.74
    & 66.00
    & 0.173
    & 3500\\
+ IVF
    & 56.28
    & 66.02
    & 0.072
    & 510 \\
+ SQ8
    & 56.14
    & 66.01
    & 0.039
    & 400 \\
\rowcolor{blue!10}
On CPU\footnotemark[2]
    & 55.53
    & 65.54
    & 0.3
    & 538 \\
\bottomrule
\end{tabular}
\end{table}
\footnotetext[1]{Brute-force search by loading the complete SNOMED CT embedding database into memory.}
\footnotetext[2]{Q8-quantized SapBERT + IVF + SQ8 indexing on CPU.}

The ontology mapping module of the Grounding Agent (Section~\ref{sec:grounding_agent}) is evaluated on the SNOMED CT Entity Linking Challenge~\cite{snomed-ct-entity-challenge} test set. As shown in Table~\ref{tab:ontology_map}, brute-force search over the complete SNOMED CT embedding space incurs the highest latency and memory usage. IVF indexing (\texttt{nlist=256}, \texttt{nprobe=64}) substantially improves efficiency with negligible impact on retrieval accuracy, while SQ8 vector quantization further reduces the memory footprint. The final deployment combines an INT8-quantized SapBERT model with IVF and SQ8 on CPU, reducing peak memory from 3500 MB to 538 MB while maintaining comparable Top-1/Top-3 accuracy, demonstrating that aggressive model and index compression can be achieved with minimal loss in ontology mapping performance.


\subsection{Concept Categorization}

\begin{figure}[!htbp]
    \centering

    \begin{subfigure}{0.24\textwidth}
        \centering
        \includegraphics[width=\linewidth]{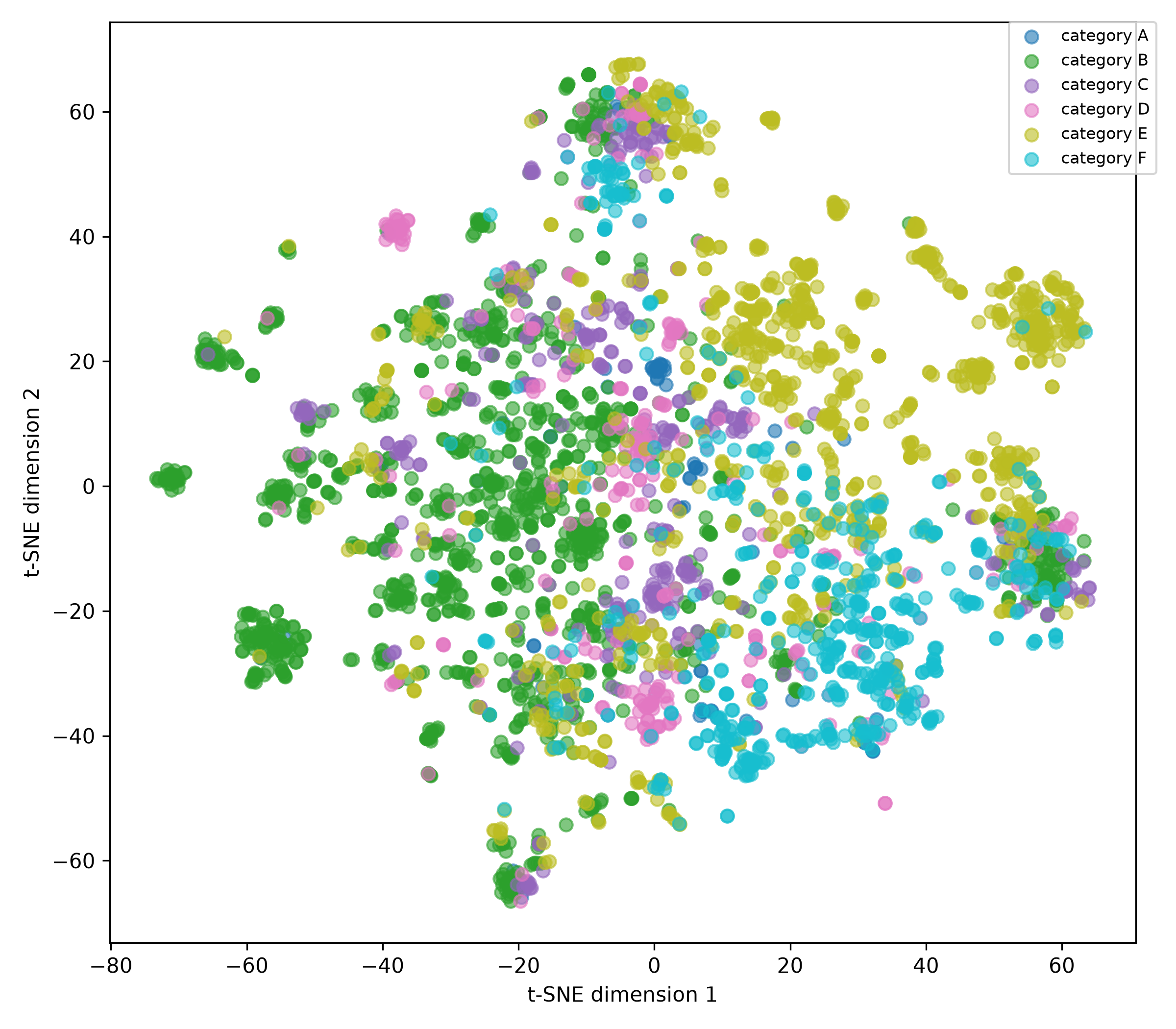}
        \caption{Base model}
    \end{subfigure}
    \hfill
    \begin{subfigure}{0.24\textwidth}
        \centering
        \includegraphics[width=\linewidth]{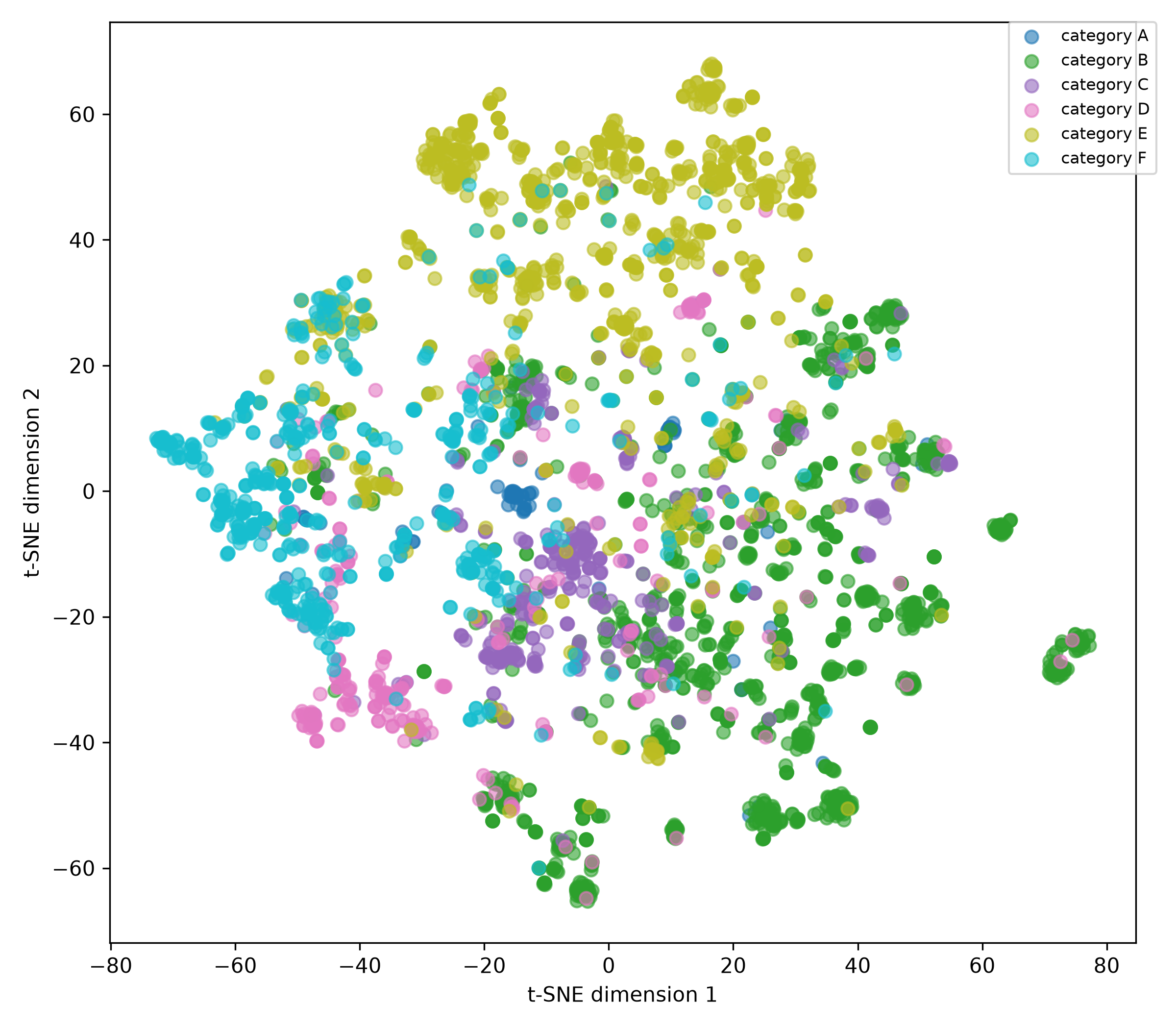}
        \caption{+SFT}
    \end{subfigure}
    \hfill
    \begin{subfigure}{0.24\textwidth}
        \centering
        \includegraphics[width=\linewidth]{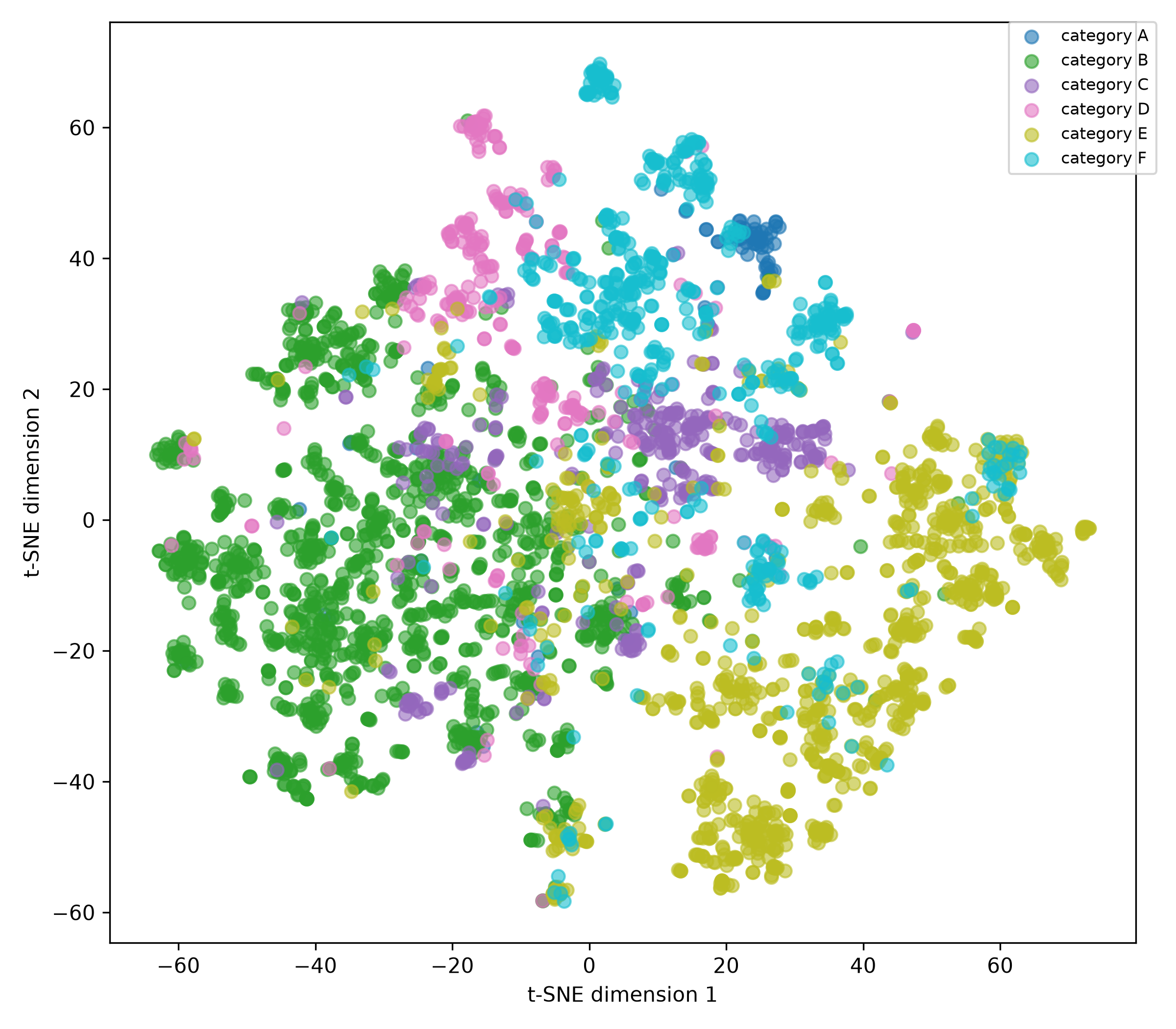}
        \caption{+KD}
    \end{subfigure}
    \hfill
    \begin{subfigure}{0.24\textwidth}
        \centering
        \includegraphics[width=\linewidth]{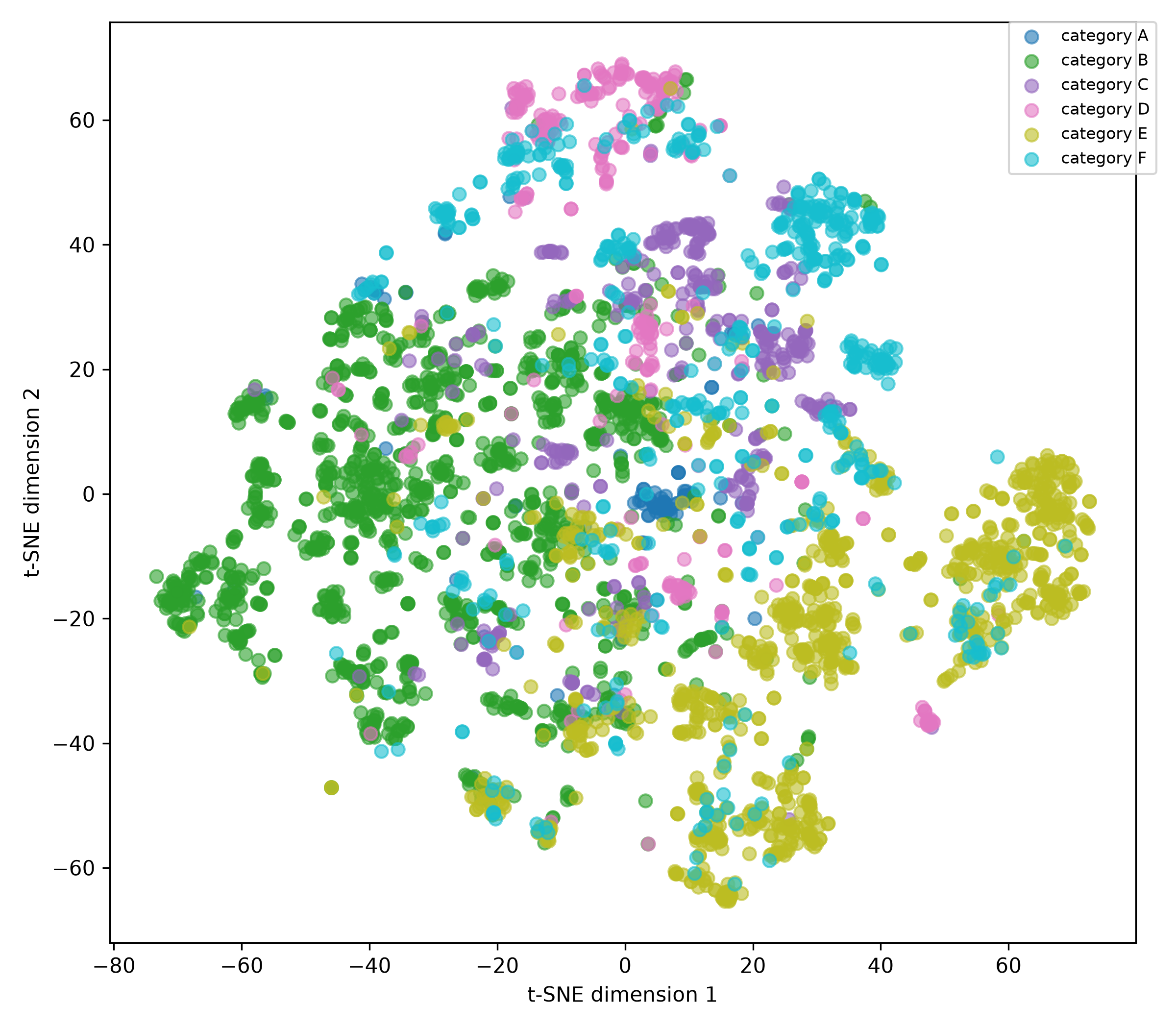}
        \caption{+IT}
    \end{subfigure}

    \caption{t-SNE plot of embedding of various models on CXR protcol categories.}
    \label{fig:tsne}
\end{figure}

\begin{figure}[ht]
    \centering

    \begin{subfigure}{0.24\textwidth}
        \centering
        \includegraphics[width=\linewidth]{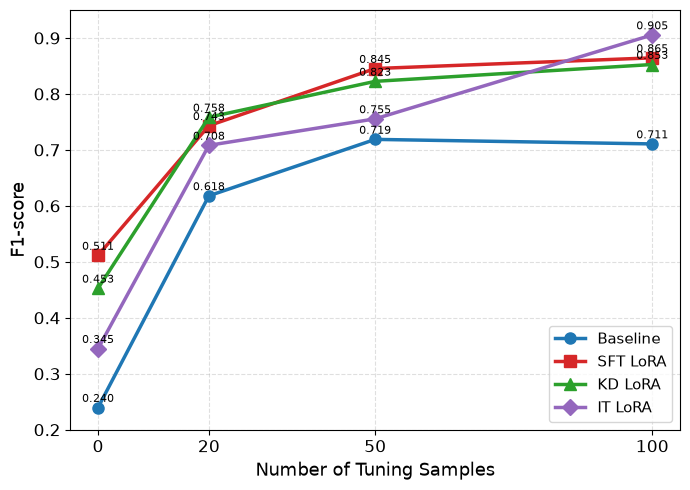}
        \caption{CXR}
    \end{subfigure}
    \hfill
    \begin{subfigure}{0.24\textwidth}
        \centering
        \includegraphics[width=\linewidth]{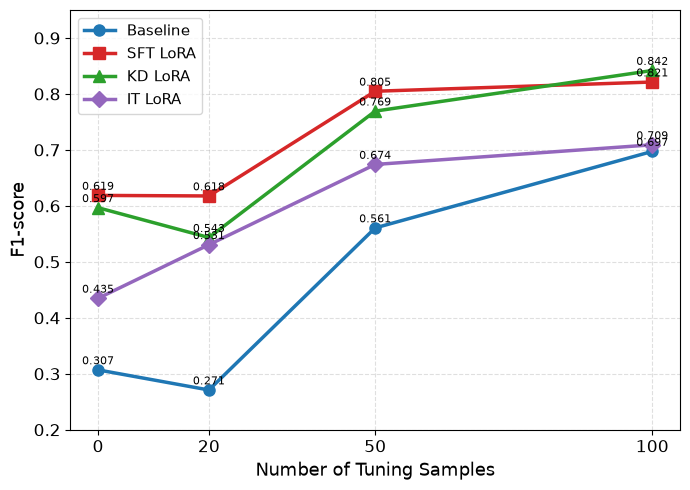}
        \caption{ChestCT}
    \end{subfigure}
    \hfill
    \begin{subfigure}{0.24\textwidth}
        \centering
        \includegraphics[width=\linewidth]{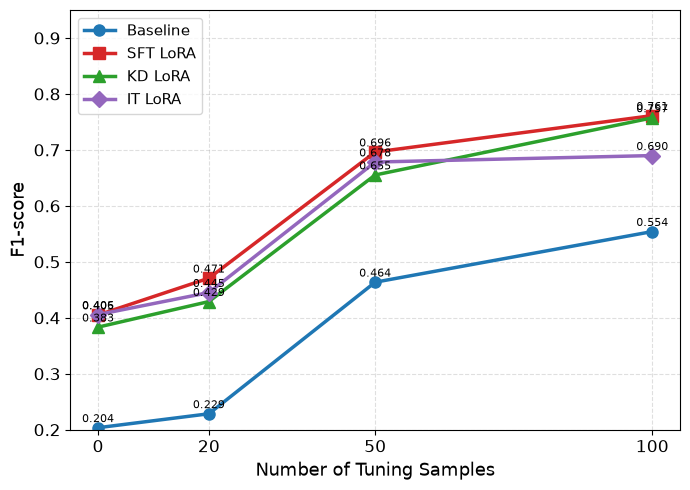}
        \caption{AbdomenCT}
    \end{subfigure}
    \hfill
    \begin{subfigure}{0.24\textwidth}
        \centering
        \includegraphics[width=\linewidth]{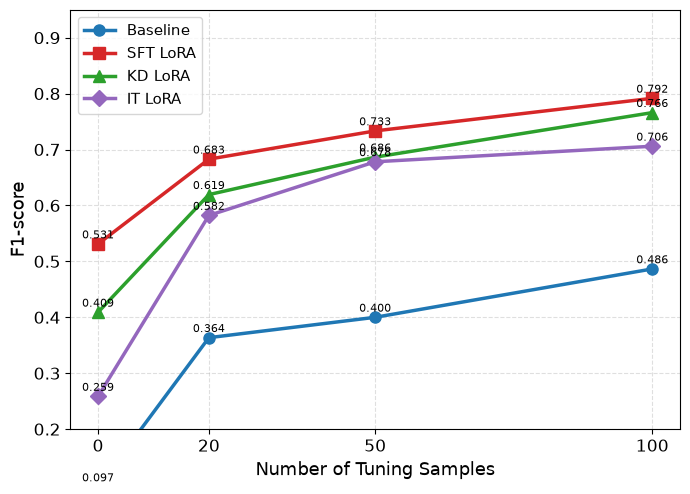}
        \caption{BrainMRI}
    \end{subfigure}

    \caption{Performance of all the models in the proposed categorization framework across various modality protocol categories vs number of finetuning samples.}
    \label{fig:categorization_f1}
\end{figure}

The proposed concept categorization framework (Section~\ref{sec:grounding_agent}) is evaluated qualitatively using t-SNE visualizations of sentence embeddings generated from enriched CXR concept--ontology text, and quantitatively on a synthetic dataset spanning the four imaging modalities. Figure~\ref{fig:tsne} compares embeddings from the base model and LoRA adapters trained at different stages (Section~\ref{sec:conc_extr}) on CXR concepts alone, showing that the combination of the base model and SFT LoRA produces the most discriminative clusters. For quantitative evaluation, 200 concept--category pairs per modality were synthesized using Claude Sonnet 5 from protocol definitions, with 100 stratified samples used for training and 100 for testing. As shown in Figure~\ref{fig:categorization_f1}, protocol prototypes learned with varying training sizes ([0, 20, 50, 100]) consistently achieve the highest F1-score when using the base model with the SFT LoRA, while the remaining LoRA adapters improve progressively with additional training data, indicating that the SFT LoRA learns more effective sentence embeddings for low-resource concept categorization.


\subsection{Performance on Edge}


\begin{table}[!htbp]
\centering
\caption{Hardware profiling of MedPAO-Assist on the targeted CPU, per stage and end-to-end under sequential vs.\ asynchronous execution. Throughput units differ by stage; all other columns share consistent units (TTFT/Total Latency in seconds, Peak RAM in GB).}
\label{tab:hardware_profile}
\fontsize{8}{9}\selectfont
\begin{tabular}{lcccc}
\toprule
\textbf{Stage}
    & \textbf{TTFT} $\downarrow$
    & \textbf{Throughput} $\uparrow$
    & \textbf{Total Latency} $\downarrow$
    & \textbf{Peak RAM} $\downarrow$\\
\midrule
MedASR                         & 0.708 & 79.425 & 18.171 & 2.68 \\
Concept Extraction             & 3.551 & 4.699  & 11.652 & 3.04 \\
Ontology Mapping               & 0.754 & 3.155  & 4.599  & 2.95 \\
Concept Categorization         & 5.355 & 0.549  & 26.352 & 3.22 \\
Structured Report Generator    & 7.947 & 4.928  & 18.545 & 3.23 \\
\midrule
\rowcolor{blue!10}
Total System (Sequential)
    & 21.722 & -- & 67.498 & 3.34 \\
\rowcolor{blue!10}
Total System (Asynchronous)
    & \textbf{4.259} & -- & \textbf{31.002} & 3.20 \\
\bottomrule
\end{tabular}

\end{table}

%



All experiments were performed on an Intel i7-1165G7 quad-core CPU with 8~GB RAM, the target deployment platform, with results in Table~\ref{tab:hardware_profile} averaged over five reports from each imaging modality. The complete pipeline requires only 3.0--3.5~GB of peak memory and achieves an average end-to-end latency of 31~s for reports averaging 260 words under asynchronous execution. Table~\ref{tab:hardware_profile} further compares sequential execution, intended for offline conversion of existing reports, with asynchronous (pipelined) execution that overlaps processing stages to support live microphone-based reporting. Throughput is reported in stage-specific units (chars/s for MedASR, tokens/s for Concept Extraction and Structured Report Generation, and concepts/s for Ontology Mapping and Concept Categorization). Peak RAM remains similar across LLM-driven stages because all models are preloaded, so per-stage values represent resident memory rather than incremental memory usage.



\section{Conclusion}
In this work, we introduced MedPAO-Assist, an edge-deployable agentic AI system that generates protocol-compliant structured radiology reports from dictation. The framework uses cooperating small language models for concept extraction, ontology grounding, and report generation, with a speech-driven human-in-the-loop correction stage to keep the radiologist in control. By coupling MedASR streaming transcription with this agentic pipeline, the system is designed to reduce manual reporting effort while preserving modality-specific protocol adherence, without relying on external APIs, subscriptions, or internet connectivity - an essential property for deployment in privacy-sensitive clinical settings. This study is an initial step: evaluation is currently limited to hardware profiling on a single target CPU, and future work will include end-to-end workflow validation with practicing radiologists, along with further compression and CPU-level optimization to reduce latency and memory usage. By enabling privacy-preserving, offline, voice-driven
structured reporting on commodity hardware, MedPAO-Assist provides a
practical pathway toward integrating agentic AI into routine radiology
workflows, particularly in settings where cloud-based solutions or
dedicated GPU infrastructure are not feasible.


    

\begin{credits}

\subsubsection{\discintname}
The authors have no competing interests to declare that are relevant to the content of this article.
\end{credits}

%

\bibliographystyle{splncs04}
\bibliography{references}

\end{document}